%% file: main.tex
\pdfoutput=1

\documentclass[11pt]{article}

\usepackage[preprint]{acl}

\usepackage{times}
\usepackage{latexsym}

\usepackage[T1]{fontenc}
\usepackage{amsmath}
\usepackage{amsfonts}
\usepackage[utf8]{inputenc}

\usepackage{enumitem}
\usepackage{microtype}

\usepackage{inconsolata}

\usepackage{graphicx}
\usepackage{booktabs}
\usepackage{xspace}

\title{Full-Duplex Dialogue Agents with Audio-Visual Understanding}

\title{Audio-video model for full-duplex speech dialogue agent}

\title{ Conversing through Noise: Audio-Visual Full-Duplex Dialogue Agents}

\title{ AV-Duplex: Audio-Visual Full-Duplex Dialogue Agents}

\title{ AV-Duplex: Audio-Visual  Full-Duplex Dialogue Assistants}

\title{AV-Duplex: Audio-Visual Full-Duplex Dialogue Amid Overlapping Speech}

\title{AV-Duplex: Audio-Visual  Full-Duplex Dialogue in Noise}

\title{AV-Duplex: Full-Duplex Dialogue Agents with Audio-Visual Input}

\title{AV-Dialog: Spoken Dialogue Models  with Audio-Visual Input}

\newcommand{\squishlist}{\begin{itemize}[itemsep=1pt,parsep=2pt,topsep=3pt,partopsep=0pt,leftmargin=0em, itemindent=1em,labelwidth=1em,labelsep=0.5em]}

\newcommand{\squishend}{\end{itemize}}

\newcommand{\xref}[1]{\S\ref{#1}}

\author{
 \textbf{Tuochao Chen\textsuperscript{1,2}},
 \textbf{Bandhav Veluri\textsuperscript{1}},
 \textbf{Hongyu Gong\textsuperscript{2}}, 
 \textbf{Shyamnath Gollakota\textsuperscript{1}}
\\
\\
 \textsuperscript{1}Paul G. Allen School of Computer Science \& Engineering, University of Washington
 \\
 \textsuperscript{2}Meta AI Research\\
 {\{tuochao,bandhav,gshyam\}@cs.washington.edu\, hygong@meta.com}
}

\begin{document}
\maketitle

\input{abs-3}

\input{intro-2}

\input{related-1}

\input{method}
\input{results-2}

\input{conclude-1}
\input{limits-1}


\bibliography{references,custom}

\appendix
\label{sec:appendix}

\section{Algorithm Latency Analysis}\label{sec:analysis}

Moshi is built on a 7B backbone model, while our unified system uses a single 8B LLaMA model. Our dual-model architecture extends this by running two such models in parallel, which naturally increases peak memory consumption.

Because system latency depends on hardware and software optimizations, which is not the focus of our paper, we focus instead on algorithmic latency which is a platform-invariant metric. In AV-Dialog, both the audio tokenizer (DAC) and the visual encoder operate causally at 25 Hz. However, the AV-HuBERT visual encoder introduces a 2-frame lookahead \cite{ma2025real}, resulting in an overall algorithmic latency of approximately 120 ms. In our dual-model setup, output tokens from the understanding module are streamed directly to the text backbone with KV-cache in parallel, enabling immediate response generation during turn-taking without additional delay.

By contrast, Moshi \cite{moshi} processes 80 ms audio chunks, achieving an algorithmic latency of about 80 ms. Note that AV-Dialog employs Moshi’s TTS model. Although our system’s latency is somewhat higher, it is mainly limited by the visual encoder’s lookahead: an aspect that could be further reduced by pretraining a visual encoder with a smaller or zero lookahead window. Despite this limitation compared to Moshi-like natively full-duplex models, we believe our approach of predicting agent's start-of-the-turn bridges the naturalness gap while also leveraging superior helpfulness and knowledge of standalone text backbones.



\section{Training Hyper-parameters}
\subsection{Stage-1 Training}\label{sec:app:stage1}

In Stage 1, we trained the original \textsc{Llama3-8b} with sequence length 4096. We use a learning rate of $3e^{-5}$ on the transformer block and a learning rate of $1.5e^{-4}$ on embedding layers and audio/visual adapters. The model is trained with 500 step warmup and trained for 50k iterations on 128 A100 GPUs with a per-gpu batch size of 1.

The proportion of each task and dataset in the stage1 training is as follows: 

\squishlist
\item \textit{Text continuation} (48.0\%): Arxiv (16.0\%), B3g (20.0\%) and Wikipedia\_en (12.0\%). 

\item \textit{Speech comprehension} (32.0\%): LibriLigh (13.76\%), MLS (13.12\%) and VP400k  (5.12\%). 

\item \textit{Audio captioning} (4.0\%): AudioSet (4.0\%). 

\item \textit{Audio-visual alignment (AVSR)} (16.0\%): Voxceleb2 (16.0\%). 
\squishend

\begin{figure*}[t!]
    \centering
    \includegraphics[width=1.6\columnwidth]{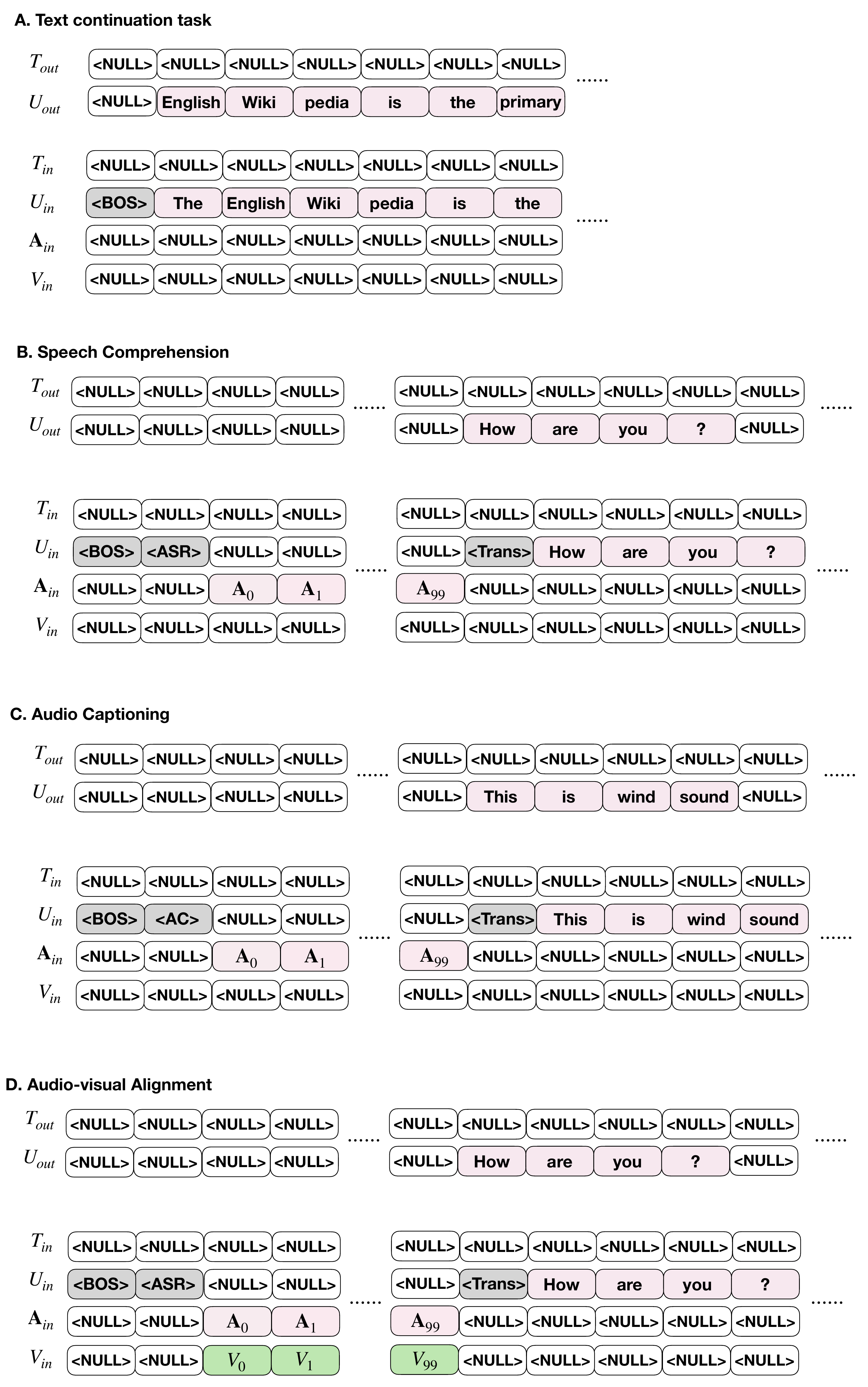}
    \vskip -0.1in
    \caption{Training input and output tokens design for different tasks at Stage 1.}
    \label{fig:appendix:stage1}
    \vskip -0.15in
\end{figure*}

The input and output token sequence design for Stage 1 training is shown in Fig.~\ref{fig:appendix:stage1}. The special tokens \verb|<ASR>|, \verb|<Trans>|, and \verb|<AC>| serve as prefixes for different tasks. We also introduce a special \verb|<NULL>| token: when a modality is missing in the input stream for a given task, it is filled with \verb|<NULL>|, whose embedding vector is all zeros after the embedding layer. If \verb|<NULL>| appears in the target stream, its loss is not computed. We apply cross-entropy loss on the output text stream.


\subsection{Stage-2 Training for Dual Model}\label{sec:app:stage2}

\begin{figure*}
    \centering
    \includegraphics[width=1.6\columnwidth]{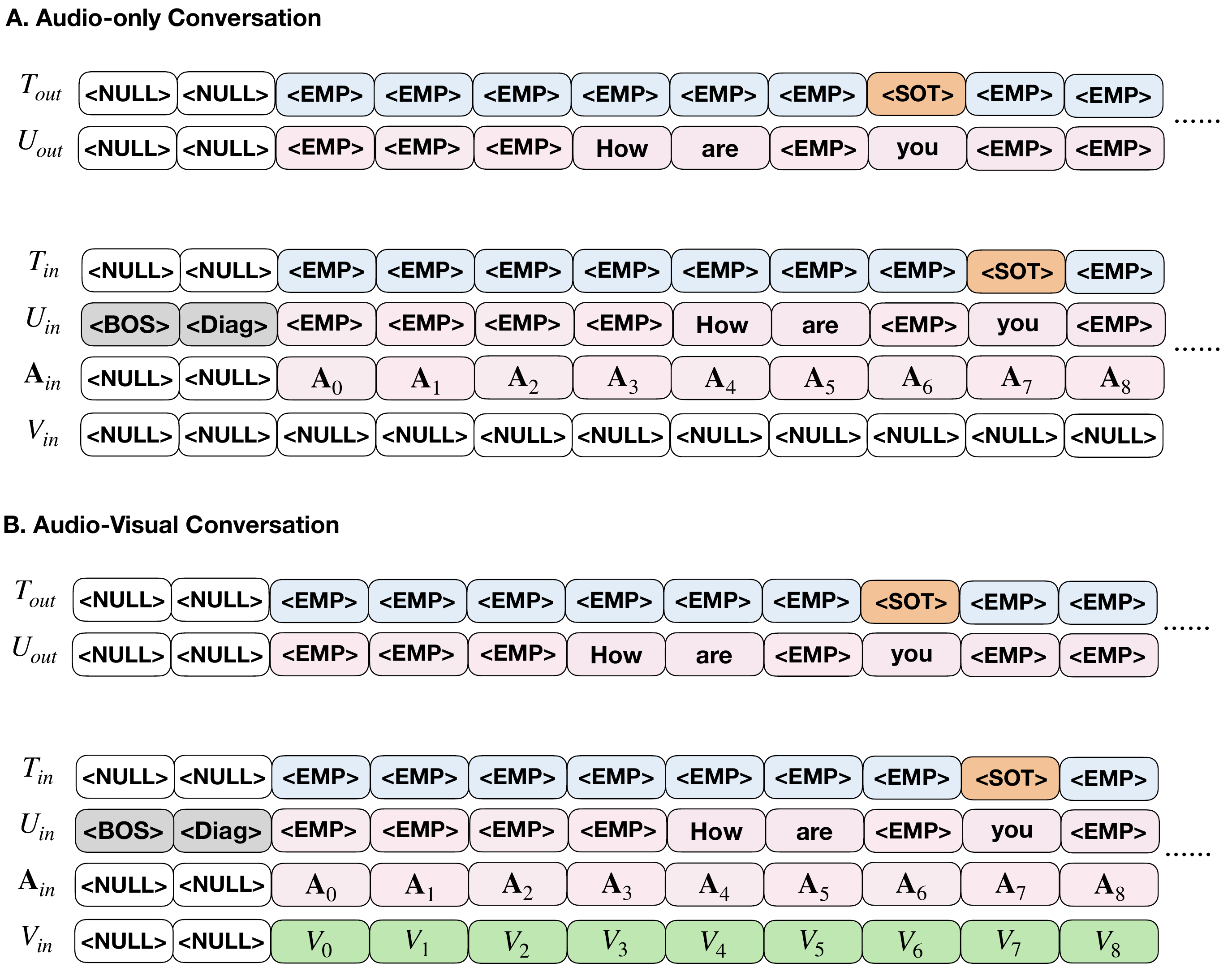}
    \vskip -0.1in
    \caption{Training input and output tokens design for different tasks at Stage 2.}
    \label{fig:appendix:stage2}
    \vskip -0.15in
\end{figure*}

In Stage 2, we fine-tuned the Stage 1 model with a sequence length of 4096. We used a  learning rate of $2e^{-5}$ for the transformer blocks, embedding layers, and audio/visual adapters. The model was trained with a 500-step warm-up over 10k iterations on 32 A100 GPUs, with a per-GPU batch size of 1. The proportion of each task and dataset in Stage 2 fine-tuning is as follows:


\squishlist
\item \textit{Audio-only conversation} (55.0\%): Fisher (10.0\%) and InterAct (45.0\%).

\item \textit{Audio-Visual conversation} (45.0\%): InterAct (45.0\%).
\squishend

The input and output token sequence design for Stage 2 training is shown in Fig.~\ref{fig:appendix:stage2}. We also apply the  \verb|<NULL>| token in the same way as Stage 1. We compute the cross-entropy loss on both the AVSR stream $U$ and Turn event stream $T$ and  compute their average. In the AVSR stream, the loss weight for text tokens is set to 1.0, while the silence token \verb|<EMP>| is set to 0.1. In the Turn-Event stream, the loss weight for Turn-taking token \verb|<SOT>| is set to 2.5, the loss weight for the backchannel token \verb|<BOT>| is set to 1.0, and the loss weight for the the silence token \verb|<EMP>| is set to 0.1.

\subsection{Stage-2 Training for Unified Model}\label{sec:app:unified}
In Stage 2 of our unified model, we fine-tuned the pretrained Stage 1 model with a sequence length of 4096. We used a  learning rate of $2e^{-5}$ for the transformer blocks, embedding layers, and audio/visual adapters. The model was trained with a 500-step warm-up over 10k iterations on 32 A100 GPUs, with a per-GPU batch size of 1.

The proportion of each task and dataset in Stage 2 fine-tuning is as follows:


\squishlist
\item \textit{Audio-only conversation} (55.0\%): Fisher (10.0\%), InterAct (45.0\%).

\item \textit{Audio-Visual conversation} (45.0\%): InterAct (45.0\%).

\squishend

\begin{figure*}
    \centering
    \includegraphics[width=1.6\columnwidth]{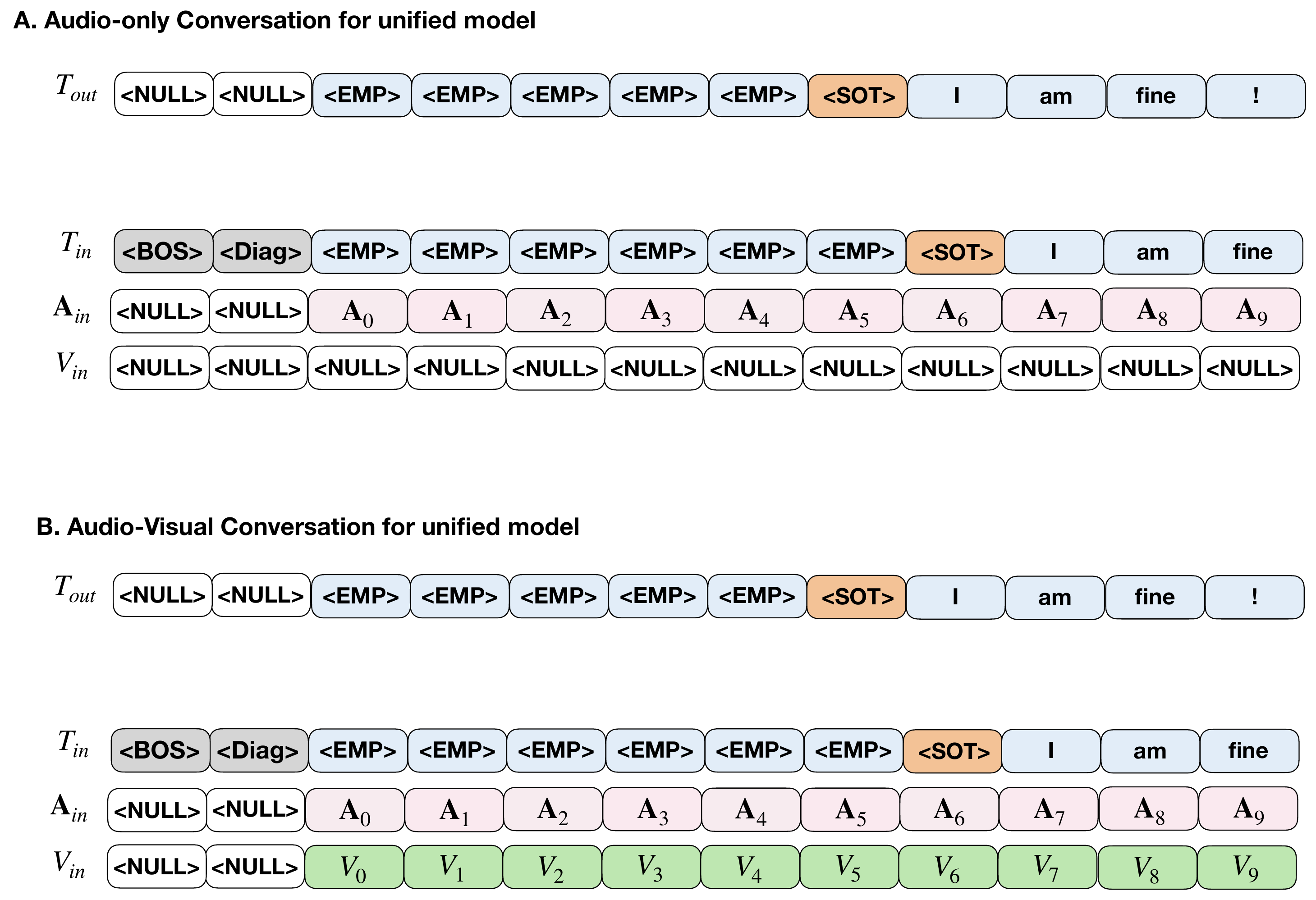}
    \vskip -0.1in
    \caption{Training input and output tokens design for multi-tasks at Stage 2 for the unified model.}
    \label{fig:appendix:unified_token}
    \vskip -0.15in
\end{figure*}

The input and output token sequence design for the unified model training at Stage 2 is shown in Fig.~\ref{fig:appendix:unified_token}. We  apply the  \verb|<NULL>| token in the same way as Stage 1. We compute the cross-entropy loss on the output stream $T$. The loss weight for text tokens is set to 1.0, the loss weight for \verb|<EMP>| is set to 0.1, the loss weight for \verb|<SOT>| is set to 2.5, and the loss weight for \verb|<BOT>| is set to 1.0.

\section{Text Backbone Hyper-parameters}
\subsection{In-Context Learning}\label{sec:texttrain}
The prompting and few-shot samples are provided as:

"Carefully read the user prompt. You follow these instructions:

You are a helpful assistant that engages in natural, casual conversation. Respond like a human would - be conversational, use natural language, and don't be overly formal. 

Here are some examples of the conversational style you should adopt:

Example1:

user: Hey, Siobhan, what's up? You seem troubled.

assistant: Yeah, I am. I'm just having a hard time. I needed someone to talk to.

user: Of course, man, I'm always here for you. What's going on?

assistant: It's just everything. Work is stressing me out. My relationship is falling apart, and I feel like I'm losing touch with my friends. I don't know what to do.

user: Well, let's start with work then. What's going on there?

assistant: It's just that... Everything is so demanding and I can't keep up. I'm constantly behind and it feels like I'm never gonna catch up.

Example2:

user: Hey, thanks for taking my motorcycle off my hands. I really appreciate it.

assistant: No problem, man. It's a nice bike. I'm sure I'll enjoy riding it.

user: I'm just glad to be rid of it, to be honest. It was costing me a lot in repairs and maintenance.

assistant: Yeah, motorcycles can be pretty expensive to keep up. They're also a lot of fun to ride. So it's worth it in my opinion.

user: I'm not so sure. I think I'd rather just... take the bus from now on. It's cheaper and less hassle.

assistant: That's a valid point. But sometimes it's nice to have the freedom that a motorcycle provides. You can go where you want, when you want.

user: I guess that's true, but it's just not worth the expense for me anymore.

Example3:
user: Would you rather have the ability to speak to the past or send messages to the future?

assistant: I would rather send messages to the future.

user: Really? But we can kind of already do that, don't you think, with certain apps and maybe hiding a letter in your house or something?

assistant: Well, to me, the progress, I like the progress, what's going on, and I can't just wait to see what else up there.

user: What kind of messages would you want to send?

assistant: And messages are positive, always positive.

Now respond to the user in this same natural conversational style. Remember your instructions."

\subsection{Instruction Tuning}\label{sec:B2}
We finetune  \textsc{Llama3-8B-Instruct} with a sequence length of  8192 on human conversational datasets, including 22.2\% InterAct dataset, 3.7\% Fisher dataset. We pre-processing the dialogue dataset by applying the  original instruction-tuning template from Llama3-8B-Instruct.
We use a low learning rate of $1e^{-5}$ and finetuned it for only 3000 steps on 32 A100 GPUs with a per-gpu batch size of 1.

\section{Evaluation Prompting for Prometheus}\label{sec:appC}
We deploy \textsc{prometheus-7b-v2.0} as our LLM evaluator on the generated response. We randomly shuffle the order of  ground-truth and generated response. The rubric description is as follow:

"Does Agent respond in a way that is generally related to the user's input and current conversation? Minor topic drift, informal language, brevity, or slight ambiguity should not be penalized. Ignore formatting, punctuation, and minor inconsistencies."

\begin{figure*}[!ht]
    \centering
    \includegraphics[width=1.6\columnwidth]{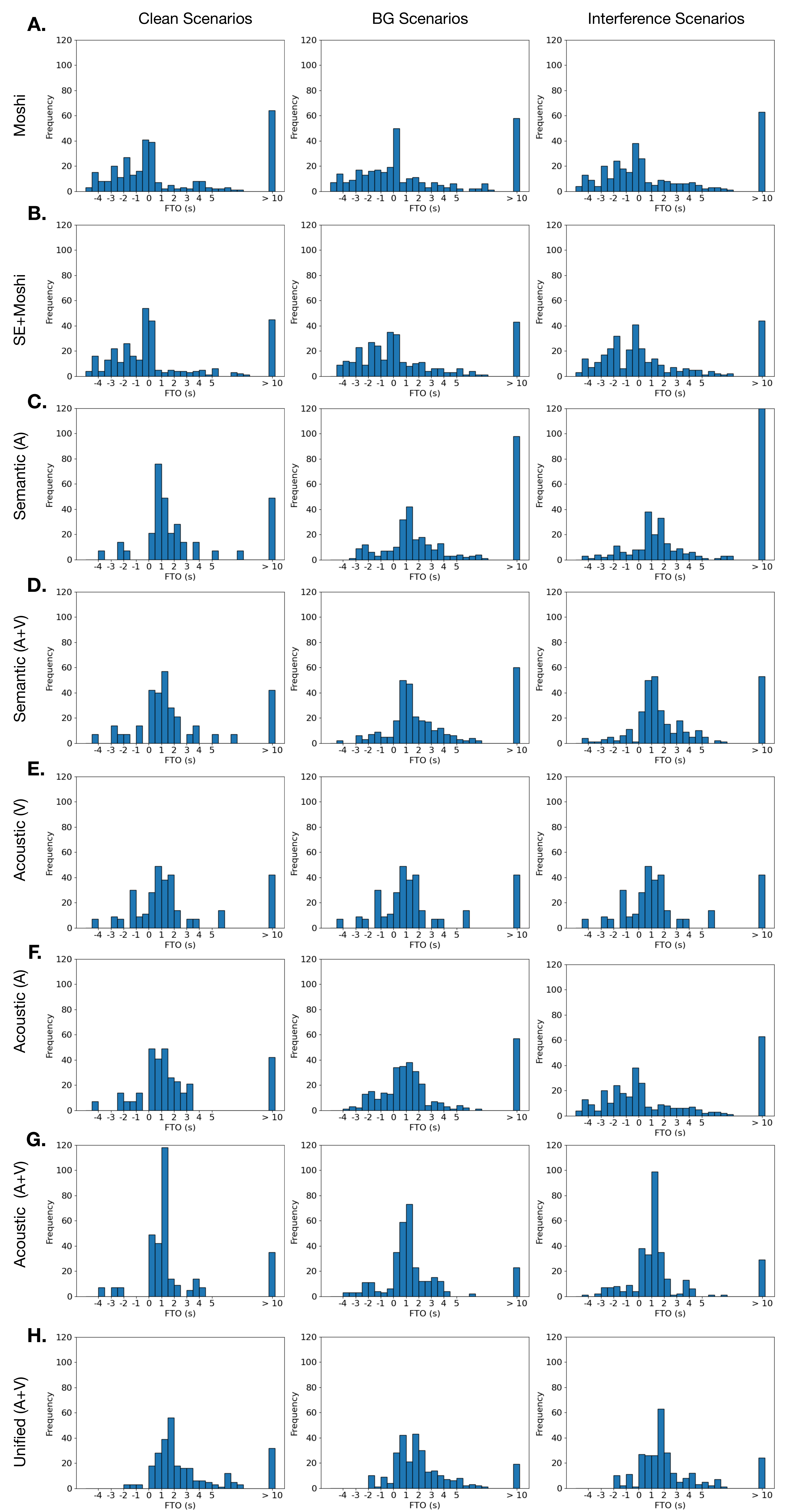}
    \caption{Distribution of FTO of different model configurations. On the x-axis of each figure, the label "> 10" also accounts for samples where the model does not respond at all.}
    \label{fig:appendix:fto}
    \vskip -0.13in
\end{figure*}

\section{FTO Distribution}\label{sec:fto}
Fig.~\ref{fig:appendix:fto} visualizes the different FTO distributions for different model configurations:
\squishlist

\item A. Moshi: state of the art speech-only dialogue model

\item B. SE+Moshi: cascade of a speech enhancement model (Demcus) and Moshi

\item C. Semantic(A): our dual-model approach with semantic tokens (DinoSR) and audio-only input

\item D. Semantic(A+V): our dual-model approach with semantic tokens (DinoSR) and audio-visual input 

\item E. Acoustic(A): our dual-model approach with acoustic tokens (DAC) and audio-only input 

\item F. Acoustic(V): our dual-model approach with acoustic tokens (DAC) and visual-only input 

\item G. Acoustic(A+V): our dual-model approach with acoustic tokens (DAC) and audio-visual input 

\item G. Unified(A+V): our unified-model approach with acoustic tokens (DAC) and audio-visual input 

\squishend

\section{N-MOS and H-MOS}\label{sec:app:scales}

{\bf N-MOS scores} for Naturalness are defined as follows: 

\squishlist
\item 1. Bad - Response is not normal English or does not make sense.

\item 2. Poor - Response is normal English but not coherent to the user's input.

\item 3. Fair - Response is somewhat plausible and coherent 

\item 4. Good - Response is plausible and coherent 

\item 5. Excellent - Response is highly plausible and coherent 
\squishend

{\bf M-MOS scores} for Meaningfulness are defined as follows: 
\squishlist

\item 1. Bad - essentially nothing in common with human-like conversation

\item 2. Poor - very little natural and human-like conversation

\item 3. Fair - substantial differences from human-like and natural conversation 

\item 4. Good - minor differences from human-like and natural conversation 

\item 5. Excellent - basically indistinguishable from human-like and natural conversation
\squishend

\begin{table}[t!]
\caption{Distribution of the samples used in human evaluation.}
\vskip -0.1in
\centering
{\footnotesize
\begin{tabular}{lccc}
\toprule
Sample Index & Noise Scenario & SNR(dB) \\
\midrule
1 & BG & 9.0 \\
2 & BG & 0.0 \\
3 & clean &   $\infty$  \\
4 & Interf & 3.0 \\
5 & clean &  $\infty$ \\
6 & Interf &  -2.99 \\
7 & Interf &  2.99 \\
8 & clean &  $\infty$ \\
9 & clean &  $\infty$ \\
10 & Interf &  -3.0 \\
11 & BG &  3.0 \\
12 & BG &  6.0 \\
13 & Interf &  -7.0 \\
14 & BG &  3.0 \\
15 & BG &  0.0 \\
\bottomrule
\end{tabular}
}
\label{tab:human_sample}
\vskip -0.15in
\end{table}

\section{Samples distribution of real-human evaluation}\label{sec:app:sampledistribution}
The properties of the samples  used in human evaluation is shown in Table.~\ref{tab:human_sample}.

\section{User study participants}
The human evaluation study was performed under our institution's IRB. All participants provided consent and were recruited from our institutions and nearby areas. 

\section{Example of AV-Dialog}\label{sec:app:example}
\begin{figure*}[!ht]
    \centering
    \includegraphics[width=2\columnwidth]{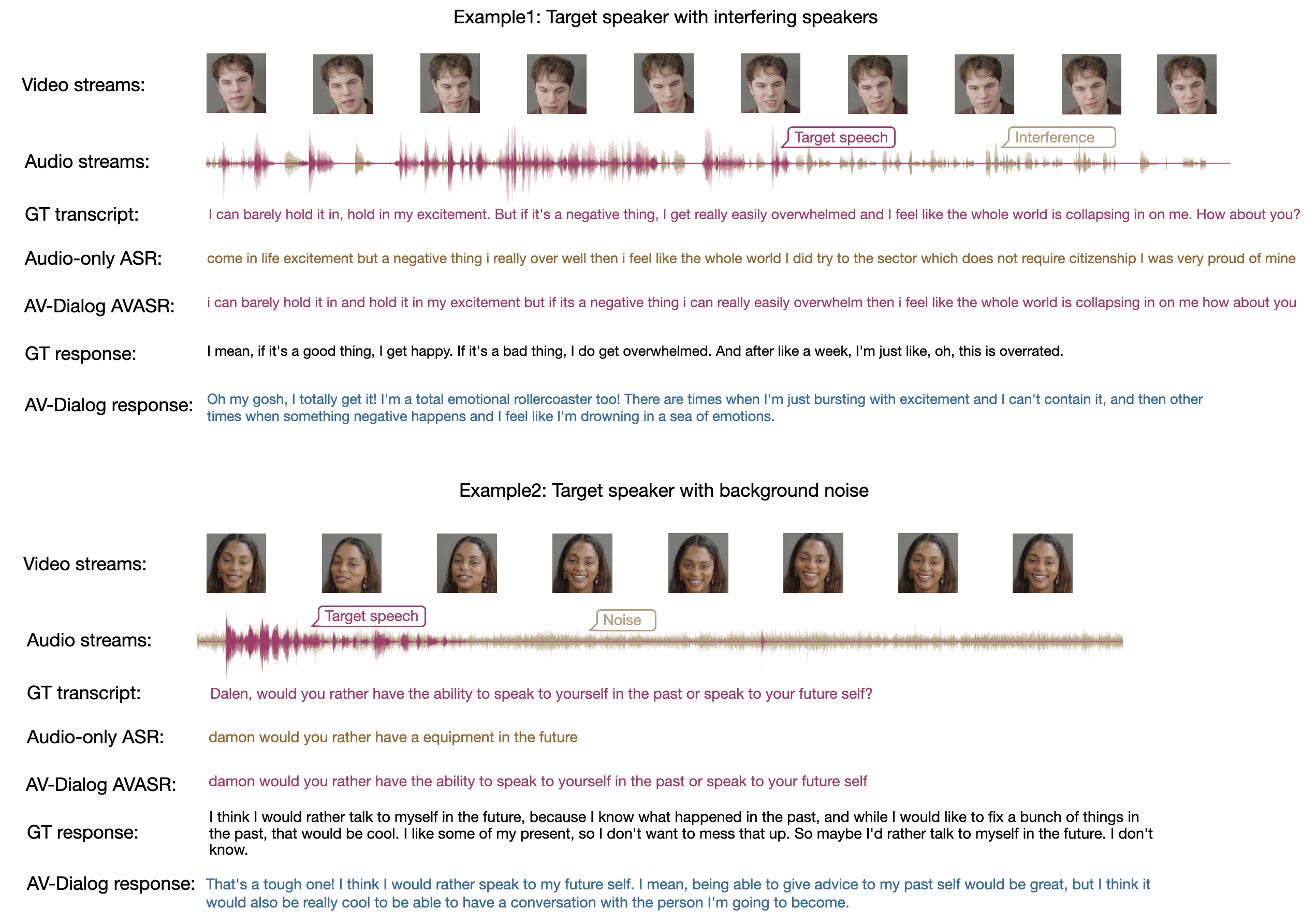}
    \caption{Data samples for our AV-dialog. Example 1 has  interfering speakers while Example 2 has significant background noise. Video and audio streams are  input  to AV-Dialog. GT transcript is the ground-truth transcription of the target speaker. Audio-only ASR is the audio-only model output. AV-Dialog AVASR is the output of AVSR stream from  AV-Dialog. GT response is the ground-truth response and   AV-Dialog response is the output of our dialogue model.}
    \label{fig:appendix:sample1}
    \vskip -0.13in
\end{figure*}

We provide some samples of our AV-Dialog input and output in Figure.~\ref{fig:appendix:sample1}

\end{document}

%% file: abs-3.tex
\begin{abstract}

Dialogue models falter in noisy, multi-speaker environments, often producing irrelevant responses and awkward turn-taking. We present AV-Dialog, the first multimodal dialog framework that uses both audio and visual cues to track the target speaker, predict turn-taking, and generate coherent responses. By combining acoustic tokenization with multi-task, multi-stage training on monadic, synthetic, and real audio-visual dialogue datasets, AV-Dialog achieves robust streaming transcription, semantically grounded turn-boundary detection and accurate responses, resulting in a natural conversational flow. Experiments show that AV-Dialog  outperforms audio-only models under interference, reducing transcription errors, improving turn-taking prediction, and enhancing human-rated dialogue quality. These results highlight the power of seeing as well as hearing  for speaker-aware interaction, paving the way for  {spoken} dialogue agents that perform {robustly} in real-world, noisy environments. \noindent {Project homepage at {\textcolor{blue}{\url{avdialog.cs.washington.edu}}}}.

\end{abstract}

%% file: intro-2.tex
\section{Introduction}
Dialogue models are moving closer to natural, human-like interaction~\cite{moshi,syncllm}, but real-world deployment remains challenging. Real environments are complex with  background noise, overlapping talk, and interfering speakers. Current models rely solely on speech inputs, making them brittle in precisely these settings; often losing track of the target speaker, producing irrelevant responses, and breaking natural turn-taking.

We argue that overcoming this limitation requires looking as well as listening. Humans  address the “cocktail party problem” by combining auditory and visual cues, using lip movements and gaze to focus on the  speaker and  learn  turn-taking cues~\cite{cocktail,doi:10.1177/23312165231152356}. 

Inspired by this, we present AV-Dialog, a novel audio-visual framework for dialogue modeling. Designing such a framework requires meeting three key challenges: First,  the model must continuously process audio and video in a streaming manner, isolating the target speaker even when background noise or louder interfering speakers are present. Second, it must detect turn-taking cues and respond appropriately, maintaining conversational flow despite overlapping or interfering  speech. Third, the system must produce coherent responses to the intended speaker without being misled by distractors or environmental noise.
large-scale text-only

\begin{figure}
    \centering
    \vskip -0.15in 
\includegraphics[width=\columnwidth]{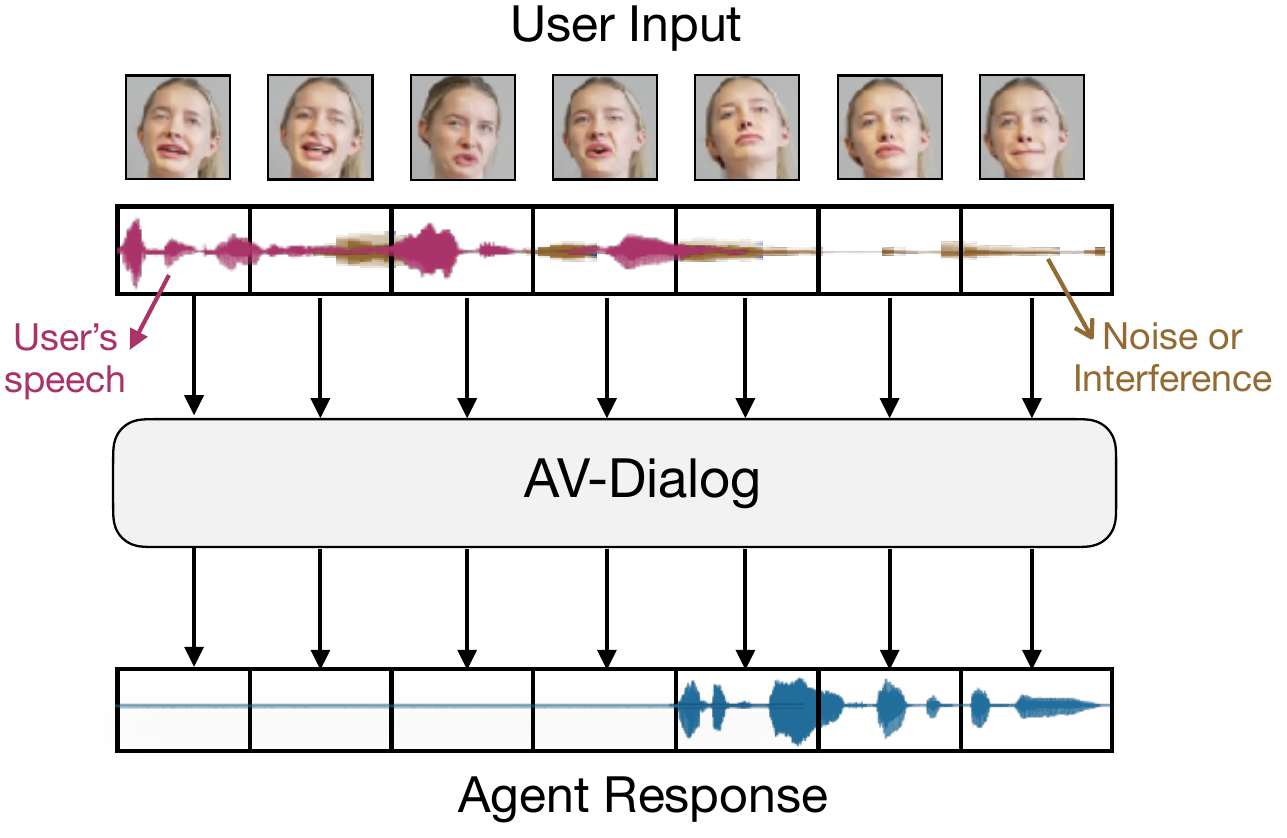}
     \vskip -0.1in
    \caption{{AV-Dialog   understands audio-visual input from the target user (purple waveform), accurately detects the appropriate time to take a turn in the conversation, and outputs responses (blue waveform), even in the presence of interfering speakers (brown waveform).} }
    \label{fig:teaser}
    \vskip -0.15in
\end{figure}

Our paper presents the first {spoken} dialog models   with audio-visual input that address the above challenges. We make the following contributions:
\squishlist
\item {\bf Multimodal dialogue modeling.}  We start with a pre-trained large language model (\textsc{Llama3-8B}) \cite{dubey2024llama} and train it to process audio and video input in a streaming manner with 40ms chunks. The model learns to extract text tokens for the target speaker under interference and predict turn-change tokens for natural conversational timing. We explore two architectures:  {\it dual} and {\it unified}. In the dual architecture, the multimodal model outputs transcriptions and turn-taking tokens that trigger a second LLM-based text backbone for high-quality response generation, either via in-context learning or instruction-tuning on dialogue data for greater naturalness. The unified architecture instead uses a single model to perform AV understanding, turn prediction, and response generation jointly. Our results show that explicit turn-change supervision is not only essential for  dual-model setups but also improves the generation quality of the unified model.


\item {\bf Acoustic tokenization for noisy, multi-speaker settings.}   Unlike prior dialogue models~\cite{moshi,syncllm} that rely on semantic tokenizers (e.g., HuBERT) trained on single-speaker speech, we use general-purpose acoustic tokens, Descript Audio Codec (DAC)~\cite{dactokens} for multimodal dialogue modeling. Because acoustic tokens preserve both semantic and raw acoustic information, they enable inherent speaker differentiation based on voice characteristics. Thus, we can better address the “cocktail party problem,” maintaining robustness to  noise and interfering speakers across a range of SNRs. In ablation studies, replacing semantic  with acoustic tokens both reduces word error rate for  streaming AVSR from 67\% to 31.7\% under strong multi-speaker interference as well as  enables more timely responses.

\item {\bf Multi-task, multi-stage training recipe.} Open  audio-visual dialogue datasets are much  smaller than text-based chat corpora, making robust training challenging. We address this with a multi-task, two-stage training strategy: the first stage trains the base LLaMA model with text prediction, ASR, AVSR and audio captioning tasks to strengthen audio-visual understanding and align with original text embeddings. The second stage fine-tunes the model on real audio-only and audio-visual conversational datasets to learn natural turn-taking and conversation context. We further improve robustness with synthetic mixture augmentation, simulating noisy, multi-speaker environments. This task-oriented approach enables AV-Dialog to acquire complementary skills from each dataset, enhancing  transcription accuracy, turn-taking prediction, and dialogue quality under challenging conditions.
\squishend


We compare AV-Dialog with Moshi-7B~\cite{moshi}, a state-of-the-art spoken dialogue model. Results show that adding the visual modality boosts turn-taking prediction accuracy from 54\% to 79\% in the presence of interfering speakers. Human evaluation (N=18) further demonstrates a +1.75-point MOS improvement in dialogue naturalness and a +1.99-point MOS gain in response relevance and helpfulness.

%% file: related-1.tex
\section{Related work}


\vskip 0.05in\noindent\textbf{Audio-visual speech recognition.}  A  related task is Audio-Visual Speech Recognition (AVSR)~\cite{whisperflamingo,watch_and_listen}, where models like AV-HuBERT~\cite{avhubert} learn speech representations from synchronized audio and video. Recent work combines pre-trained audio~\cite{whisper} and video~\cite{avhubert}  with language models to improve word error rates~\cite{10889251}. While AVSR systems excel at speech recognition, they are not designed for generative dialogue, turn-taking, or full-duplex interaction. In contrast, AV-Dialog extends audio-visual fusion beyond recognition to enable grounded, interactive conversational agents. Moreover, most AVSR models operate offline with full-recording access~\cite{whisperflamingo,10889251}, whereas our system performs streaming inference and incorporates AVSR as one of its multi-task objectives. We therefore compare AV-Dialog with state-of-the-art streaming AVSR models such as Auto-AVSR~\cite{ma2023auto} in~\xref{sec:avsrcomp}.

\vskip 0.05in\noindent\textbf{Dialog models.} Recent work on dialog models  generate spoken responses from a given prompt. Notably, SpeechGPT~\cite{speechgpt} is finetuned on speech-only data and multimodal instructions for spoken question answering. Multi-modal models like SpiritLM~\cite{nguyen2024spiritlm} accept either speech or text as prompts and generate responses in either modality, while prior non-open source models~\cite{realtalk} handle audio-visual inputs  but require explicit prompting. Unlike AV-Dialog, these systems  do not model turn-taking and hence do not know when to respond, which is a key component of human-like dialog interaction.


Recent full-duplex dialogue models like 
dGSLM~\cite{lakhotia2021generative}, Moshi~\cite{moshi}, and SyncLLM~\cite{syncllm} generate responses concurrently with user input by predicting intent or turn-endings without explicit prompts. However, relying solely on text and semantic speech tokens limits their ability to track the target speaker in noisy, multi-speaker settings. In contrast, AV-Dialog integrates visual cues and general-purpose acoustic tokens for robust speaker tracking and dialogue generation under challenging signal-to-noise (SNR) conditions.

%% file: method.tex
\begin{figure*}[!ht]
    \centering
    \includegraphics[width=2\columnwidth]{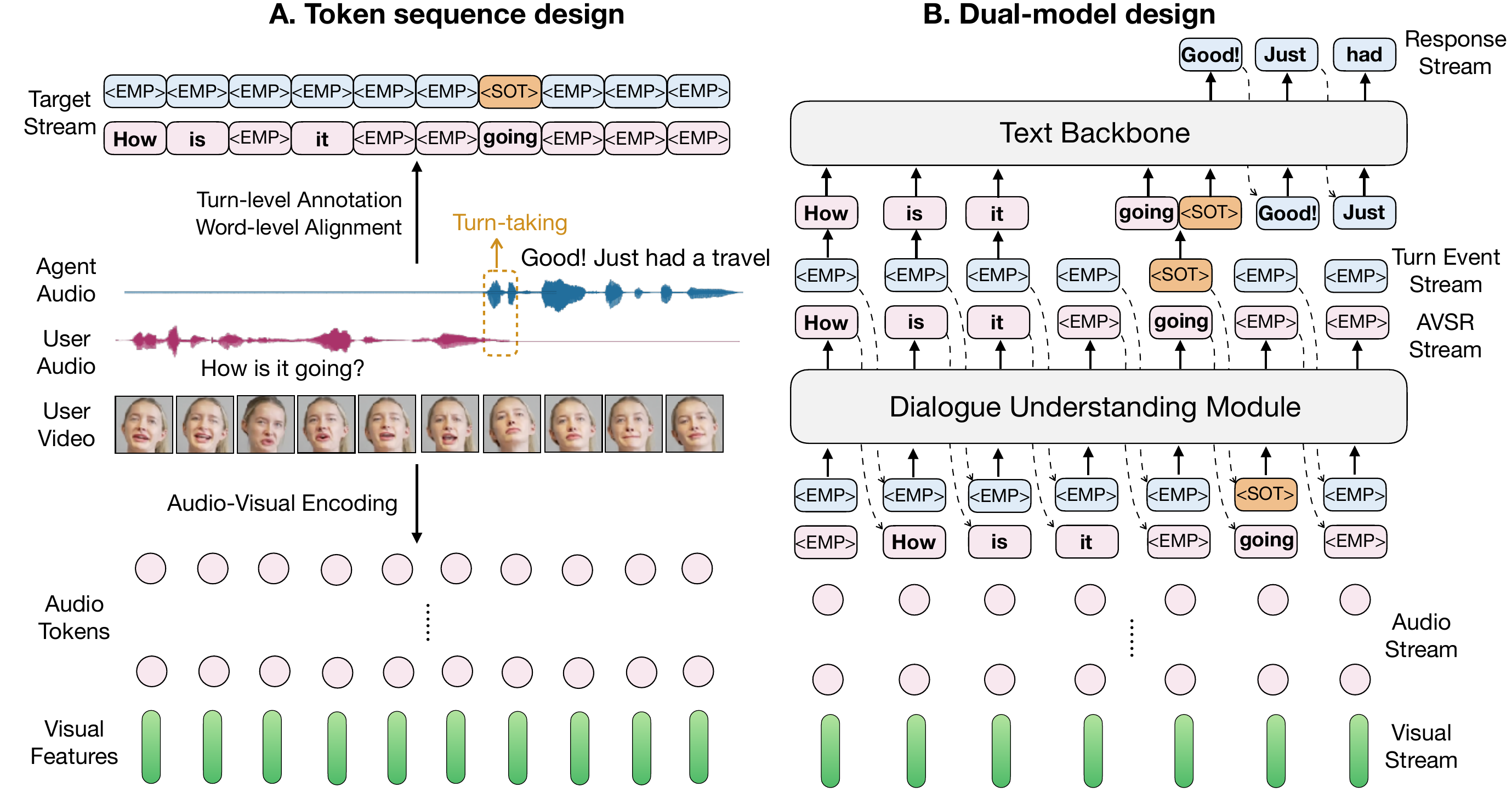}
    \vskip -0.1in
    \caption{{Token sequence and dual-model design. {\bf A.}  We use a  DAC tokenizer to encode  audio into 16 audio token streams and use AV-HuBERT to convert video to continuous visual features. We use turn-level annotation and word-level alignment to generate the target output text stream. {\bf B.} shows our dual-model pipeline for AV-dialog. The AV dialogue understanding module recognizes user speech and detects potential turn-taking events, while the text backbone generates high-quality responses once turn-taking is triggered.}}
    \label{fig:model}
    \vskip -0.15in
\end{figure*}

\section{AV-Dialog Models}

In human conversation, we process rich acoustic and visual cues to understand and respond via speech.  Prevalent  dialogue models, however, emphasize the modality the agent must generate (speech \& language) while only considering the semantic representations of speech and/or ignore visual cues; making them brittle to noise and interference.  In contrast, combining audio and visual context enables accurate turn-boundary detection and timely responses. To this end, we develop a dialogue framework built on  audio-visual understanding that infers the user’s complete intent, both what is said and when they intend to yield the floor.

As shown in Fig.~\ref{fig:model}B, AV-Dialog's dual-model architecture comprises two components: an audio-visual dialogue understanding module and a text backbone. The latter is implemented with instruction-tuned \textsc{Llama3-8B} \cite{dubey2024llama}, though any text LLM or API can be used.


\subsection{Audio-Visual Dialogue  Understanding}

We propose an AV dialogue understanding model, fine-tuned on a base text-LLM, with two essential capabilities for  voice interfaces: recognizing user speech and detecting intent to yield the conversation floor. It must also operate as a streaming model, so we can update the response LLM’s KV-cache as the user speaks.


Our model processes multi-stream inputs: at each timestep $n$, a continuous visual stream $V_{n}$ from a visual encoder and 16 audio streams $\mathbf{A}_n = [A_{n, 1}, \dots, A_{n, 16}]$ from an audio tokenizer. It outputs two streams: (1) a text stream $U_n$ representing the user’s input, and (2) a turn event stream $T_n$ from the AV understanding module, predicting when the agent should take the conversation floor.



Both streams are synchronized and operate on 40 ms chunks.  Both embeddings are projected to the transformer’s model dimension via separate linear layers and summed with the previous timestep’s text embedding to produce the final  embedding $e_n=\mathcal{L}_A(\sum_{i=1}^{16} \mathcal{E}(A_{n, i})) + \mathcal{L}_V(V_n) 
 + \mathcal{E}(U_{n-1}) + \mathcal{E}(T_{n-1})$. 
Here, $\mathcal{E}(\cdot)$ denotes  the embedding layer, $\mathcal{L}_V(\cdot)$  the visual projection layer, and $\mathcal{L}_A(\cdot)$  the audio projection layer. 

At the AV-Dialog model output $z_n$, two linear heads estimate the distributions of  $U_n$ and $T_n$, conditioned on all preceding sub-sequences.
\begin{align*}
&  \boldsymbol{\sigma}(L_{U}(z_n)) \approx  \mathbb{P}[U_n | \mathbf{A}_{\leq n}, V_{\leq n}, U_{< n}, T_{<n}], \\
&  \boldsymbol{\sigma}(L_{T}(z_n)) \approx  \mathbb{P}[T_n | \mathbf{A}_{\leq n}, V_{\leq n}, U_{< n}, T_{<n}]  
\end{align*}
$\boldsymbol{\sigma}(\cdot)$ is the softmax operation, $L_{U}(\cdot)$ the linear header for the text stream, and $L_{T}(\cdot)$ the linear header for the turn event stream.



\subsubsection{Audio-Visual Encoding}

Most prior turn-taking and spoken dialogue models rely on speaker-invariant semantic speech representations~\cite{dgslm, syncllm, moshi}. While effective in clean settings, they struggle in real-world, “cocktail-party” environments. Models like HuBERT~\cite{hsu2021hubert}, though robust to uncorrelated background noise, can amplify spurious speech interference due to their speaker invariance.

We instead leverage general audio representations, enabling inherent speaker differentiation based on voice characteristics. Our AV-Dialog model uses the high-fidelity Descript Audio Codec (DAC)~\cite{dactokens} tokenizer, where each 40 ms  chunk is encoded into 16 DAC  codebooks.

We incorporate visual cues of the speaker because they (1) enable robust target speech identification in noisy environments, (2) enhance speech perception and understanding~\cite{avhubert, cappellazzo2025large}, and (3) provide crucial signals for estimating turn boundaries.  
Specifically, we use the dlib library~\cite{dlib} to detect face regions in first-person video and extract continuous lip-centric visual representations via a pre-trained AV-HuBERT model~\cite{avhubert}.

\subsubsection{Output Streams}
\label{output_seqs}

The AV dialogue understanding module outputs  two token streams: i) time-aligned transcription of user's speech $U_n$, and ii) turn-taking event labels $T_n$, using the $L_U$ and $L_T$  heads, respectively.


If a user’s word begins at $t_{start}$, the model predicts its tokens from timestep $\lceil t_{start}/25 \rceil + d$, where $d$ is a small delay providing a reasonable context for recognizing the word. When no word is uttered, it outputs a silence token \verb|<EMP>|.


For turn boundaries, we adopt PairwiseTurnGPT’s~\cite{leishman2024pairwiseturngpt} turn-taking event  taxonomy: (1) \verb|Normal turn|, agent speaks after the user finishes; (2) \verb|Overlapping turn|, agent begins before user finishes, i.e, partial overlap; (3) \verb|Backchannel|, short interjections (e.g., “hmm”, “yeah”). The model predicts special token \verb|<SOT>| for both \verb|Normal| and \verb|Overlapping| turns, and \verb|<SOB>| for \verb|Backchannels|. For timesteps  without turn taking events, we output \verb|<EMP>|. An example turn-taking event stream following this schema is  in Fig~\ref{fig:model}A.


\subsection{Response Generation}

Our audio-visual dialogue model, in its dual-model mode,  streams user speech recognition and turn-taking predictions directly into a text-based LLM backbone (Fig.~\ref{fig:model}B). This design combines the responsiveness of instruction-tuned LLMs with the flexibility to swap in different text LLMs or APIs.

The text backbone operates in two states: \textit{LISTENING} and \textit{SPEAKING}. In \textit{LISTENING}, non-silence tokens from the AV understanding module are streamed as the user’s input. When a turn-taking token appears, the model switches to \textit{SPEAKING}, generating responses autoregressively. If  new user speech tokens arrive mid-response, the model yields the floor and re-enters \textit{LISTENING}.




To make  response more natural and human-like, we explore two methods:
\squishlist

\item \textbf{In-Context Learning (ICL):} We apply in-context learning~\cite{icl} and add few-shot dialogue examples from \textsc{Seamless Interaction} (InterAct) \cite{seamlessinteraction} training sets to the text backbone's system prompt (see ~\xref{sec:texttrain}.1).

\item \textbf{Instruction Tuning (IT):} We finetune a chat-oriented LLM on real human dialogues using instruction tuning~\cite{instructiontuning} to improve naturalness and responsiveness. The finetune hyperparameters can be found in (see ~\xref{sec:texttrain}.2).


\squishend

The generated text is then converted to speech via the streaming TTS module Mimi~\cite{moshi}.

\subsection{Training Strategy}

A key challenge in training our AV-Dialog understanding module is the scarcity of large-scale aligned audio-visual conversational data. To address this, we leverage diverse dataset sources: text dataset, monadic audio/audio-visual datasets, and real dyadic audio-only and audio-visual conversational data.
We build on a pre-trained text LLM, \textsc{LlaMa-3-8B}, and employ a two-stage, multi-task training approach to progressively develop the audio-visual understanding needed for a robust dialogue model.



\subsubsection{Stage 1: Audio-Visual Understanding} 

The first stage focuses on aligning text, audio, and visual modalities through four multi-task objectives (see~\xref{sec:app:stage1} for training hyper-parameters):
\squishlist
\item {\it Text continuation:} Utilize large-scale text-only datasets for text continuation pre-training objective, to preserve robust language understanding and avoid catastrophic forgetting of textual data.  
\item  {\it Speech comprehension:} Train on monaural speech datasets,  LibriLight \cite{librilight}, MLS \cite{Pratap2020MLSAL}, and VP400k \cite{wang-etal-2021-voxpopuli}, on the ASR task to provide the model with  acoustic comprehension of human speech.

\item {\it Audio captioning:} Use the large audio dataset, Audioset~\cite{audioset}, for audio captioning task to achieve general audio comprehension. 

\item {\it Audio-visual alignment:} Train AVSR task using  VoxCeleb2~ \cite{DBLP:journals/corr/NagraniCZ17}, which is a large audio-visual monadic dataset. This enables the model to learn visual features linked to speech, fostering multimodal learning across text, audio, and visual modalities. To further improve AV understanding in noisy conditions, we apply the synthetic mixing augmentation from \xref{mix} on this dataset.

\squishend




\subsubsection{Stage 2: Learning about  Conversations}

We train the model on audio-only and audio-visual conversational data to learn  natural dialogue dynamics. We use Fisher~\cite{Cieri2004TheFC} for audio-only and InterAct~\cite{seamlessinteraction} for audio-visual conversations, optimizing two tasks: (1) streaming AVSR and (2) turn-taking event prediction. Synthetic mixing augmentation (\xref{mix}) is also applied for  robustness in noisy settings. Training hyperparameters are detailed in~\xref{sec:app:stage2}.



{To prepare target sequences, we align words and turns by converting conversations into synchronized token streams. We deploy Whisper-Large~\cite{radford2022robust} to acquire word-level timestamps. The first token of each word is placed at the $\lceil t_{start}/25\rceil + d$ token  in the AVSR stream, where $t_{start}$ is the start timestamp from Whisper.  The special turn event token is placed at $\lfloor t_{turn}/25\rfloor$, where $t_{turn}$ is the timestamp of the annotated turn event. Note that, the $d=1s$ is also added to the AVSR stream but not to  the Turn event stream to avoid introducing additional delays to the  response.}


\subsubsection{Synthetic Mixing Augmentation}\label{mix}

We apply synthetic mixing to simulate noisy, multi-speaker environments. For each training sample, with  20\% probability, we  use clean audio as input, with 40\% probability, we  mix the clean audio with background noise randomly sampled from from MUSAN~\cite{snyder2015musan}, and with 
40\% probability, we mix with 1–4 interference speakers from the same dataset. The input SNR is uniformly sampled between –8 dB and 8 dB. This augmentation enables the model to understand audio-visual cues in complex, real-world conditions.


\begin{figure}
    \centering
    \includegraphics[width=0.8\columnwidth]{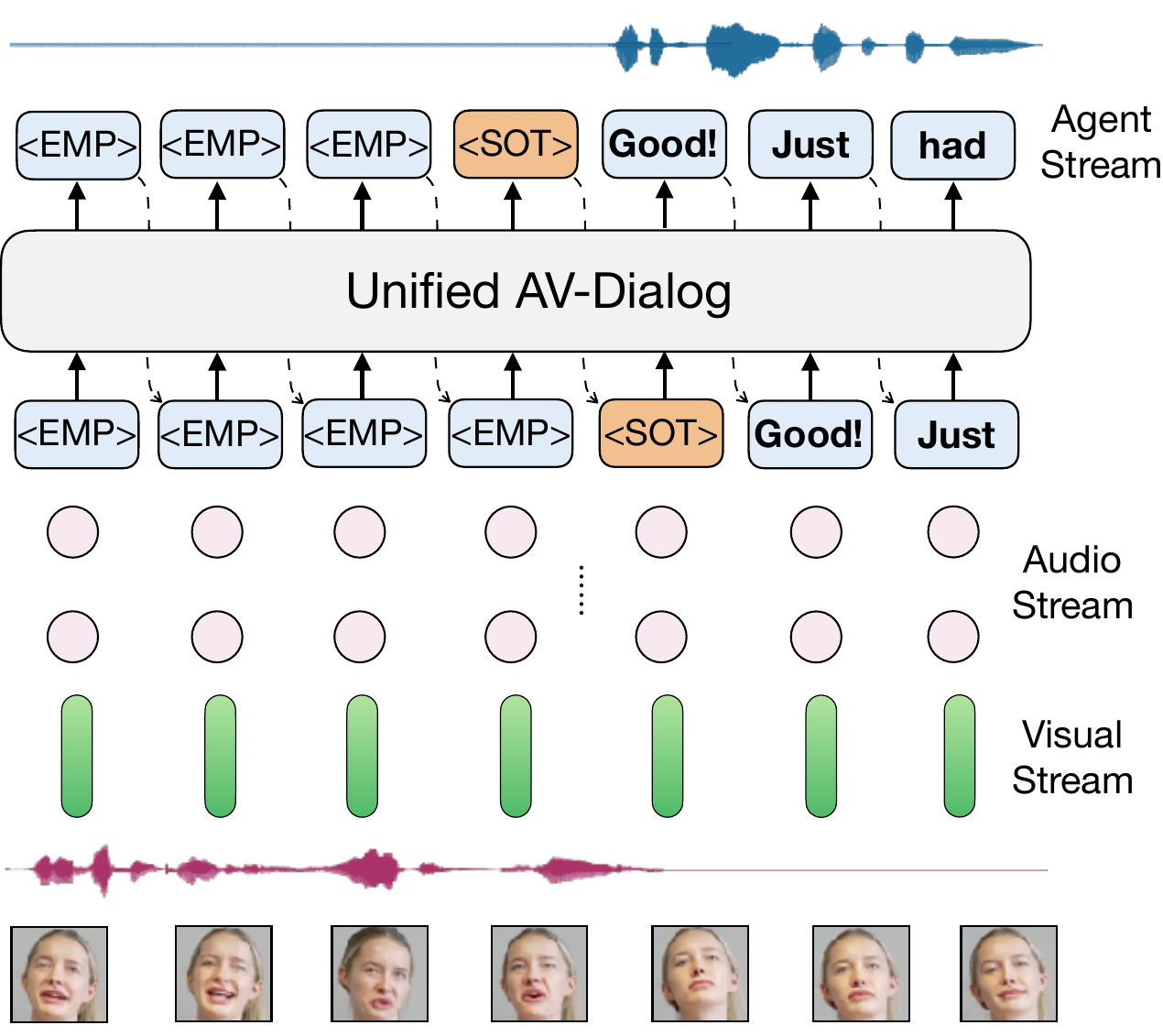}
    \caption{Unified AV-Dialog model. It takes the audio-visual input and predicts the turn-taking events. When the special turn-taking token is generated, the AV-Dialog model  generates the response on the same output stream. }
    \label{fig:unified}
    \vskip -0.13in
\end{figure}

\subsection{Unified AV-Dialog Model}
We also  explore a unified model variant where the AV-Dialog module directly generates full-duplex responses, eliminating the need for a text backbone (see~\xref{sec:analysis} for algorithmic latency analysis). This requires two key modifications:
\squishlist
\item {\it Model and token design:} We remove the AVSR stream and instead add time-aligned agent response tokens interleaved with the turn-taking event stream (Fig.~\ref{fig:unified}). Empirically, we found removal of AVSR stream  improves unified model performance, enabling it to predict turn-taking events and then generate responses directly.


\item {\it Training setup:}  In Stage 2, we train on the Fisher and InterAct datasets to generate aligned text responses alongside turn-taking predictions (Fig.~\ref{fig:unified}). Training hyperparameters are provided in~\xref{sec:app:unified}.


\squishend

%% file: results-2.tex
\section{Experiments}



For each AV dialog sample, one side is randomly chosen as the user, whose audio and visual tokens are streamed into the model while output tokens are generated simultaneously. We focus on three aspects: (1) how well the model understands audio-visual input, (2) how accurately it predicts turn-taking, and (3) the quality of its responses. 

To test robustness in noisy conditions, we evaluate under three conditions:

\squishlist
\item \textit{Clean:} Clean raw audio as input. 

\item \textit{BG:} Clean raw audio mixed with background noise (music, chatter) from MUSAN~\cite{snyder2015musan}, input SNR range is -8dB to 12dB.


\item  \textit{Interf:} Clean raw  audio mixed with 1-4 interfering speakers from the same dataset as the target speaker, input SNR range is -8dB to 12dB.

\squishend









\subsection{Audio-Visual Understanding Evaluation}\label{sec:avsrcomp}
We evaluate streaming AVSR  using word error rate (WER) as the metric. We  compare our model with the state-of-the-art streaming AVSR model, Auto-AVSR~\cite{ma2023auto}. More recent AVSR works~\cite{whisperflamingo,10889251} focus on offline settings, where models process the full recording before inference, which is an easier task. So, we benchmark against Auto-AVSR on Voxceleb2 datasets.\footnote{We do not report results on LRS3, as the dataset is no longer publicly available.}

For a fair comparison, we first benchmark on the Voxceleb2 test set, as both our model and Auto-AVSR are trained on its training set. Since Voxceleb2 lacks text labels, we use Whisper-Large to transcribe clean speech as ground truth for WER. Table.~\ref{tab:vox2} compares audio-only, video-only, and audio-visual models. AV-Dialog  achieves consistently lower WER than Auto-AVSR, with audio-visual input yielding the best performance.

\begin{table}[t!]
\caption{Benchmarking streaming AVSR on the test set of Voxceleb2. We measure WER (\%) on the four model: 1) Auto-AVSR (A+V): state-of-the-art of streaming AVSR model 2) Ours (A): our model trained and test on audio-only input. 3) Ours (V): our model trained and tested on visual-only input. 4) Ours (A+V): our model trained and tested on audio-visual input.}
\vskip -0.1in
\centering
{\footnotesize
\begin{tabular}{lcccc}
\toprule
WER(\%) $\downarrow$ & Clean & BG & Interf \\
\midrule
Auto-AVSR (A+V) & 26.8 & 48.2 & 71.8 \\
Ours (A) & 18.0 & 60.2 & 76.3 \\
Ours (V) & 87.1 & 87.1 & 87.1\\
Ours (A+V) & \textbf{17.4} & \textbf{35.6} & \textbf{38.8} \\
\bottomrule
\end{tabular}
}
\label{tab:vox2}
\vskip -0.15in
\end{table}

\begin{table}[t!]
\caption{Streaming AVSR  on the InterAct test set. } 
\vskip -0.1in
\centering
{\footnotesize
\begin{tabular}{lcccc}
\toprule
WER(\%)$\downarrow$ & Clean & BG & Interf \\
\midrule
Ours (A) & 28.6 & 68.0 & 92.2 \\
Ours (V) & 67.8 & 67.8 & 67.8\\
Ours (A+V) & 16.3 & 37.4 & 30.8 \\
\bottomrule
\end{tabular}
}
\label{tab:avsr}
\vskip -0.15in
\end{table}

\begin{figure}
    \centering
    \includegraphics[width=\columnwidth]{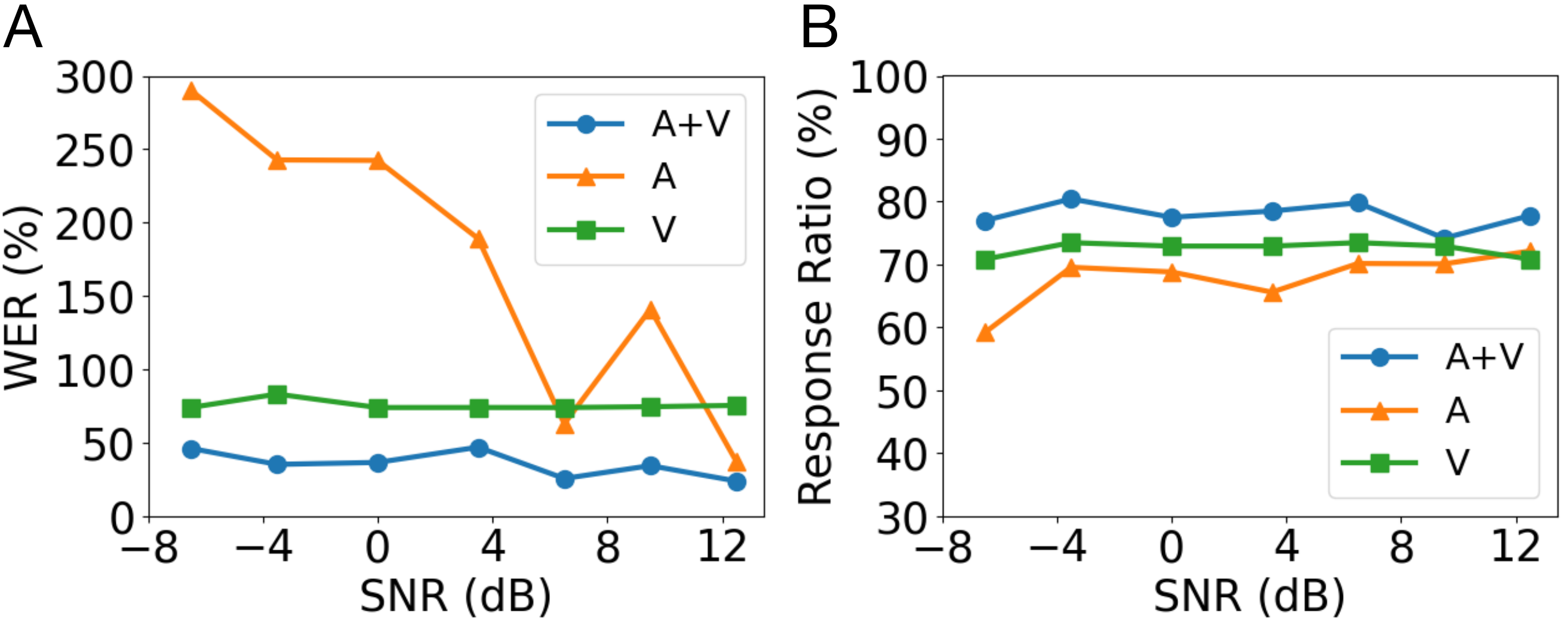}
    \caption{Model performance across different SNRs. Plot A shows the WER of streaming AVSR task on different SNRs of noisy audio input. Plot B shows the response ratio of the turn-taking prediction on different SNRs of noisy audio input.}
    \label{fig:snr}
    \vskip -0.13in
\end{figure}

We also evaluate our models for the  streaming AVSR task on the test-set of the InterAct dataset. We use the transcription from InterAct as our ground-truth text to compute WER.  As shown in Table.~\ref{tab:avsr}, our AV-dialogue model achieves much lower WER than audio-only or visual-only input, demonstrating that combining modalities greatly improves AV understanding, especially under noise and interference. To further assess robustness, we evaluate across SNR ranges (Fig.~\ref{fig:snr}A), averaging WER over multiple noise samples for BG and Interf scenarios. The results show our AV  model remains robust across varying SNR levels.



\begin{table*}[!t]
\caption{Turn-taking evaluation under different noise and interference conditions.}
\vskip -0.1in
\centering
{\footnotesize
\begin{tabular}{lcccccccccc}
\toprule
Model &  &Clean&  & &BG &  & &Interf & \\
\midrule
Metrics & Reponse    & FTO & Median & Reponse   & FTO & Median  &  Reponse    & FTO  & Median  \\
&  Ratio($\uparrow$)  & Err($\downarrow$) &  FTO &  Ratio($\uparrow$)  &  Err($\downarrow$) &  FTO &  Ratio ($\uparrow$) &  Err($\downarrow$)  & FTO \\

\midrule
Moshi & 54.0\%  & 3.48 & -0.72 & 53.8\%  & 3.20 & -0.12 & 52.5\%  & 3.27 & -0.52 \\
SE + Moshi & 55.9\% & 3.66  & -0.6 & 56.0\% & 3.24 & -0.7  & 54.4\% & 3.36  & -0.84 \\
Ours (A) & 73.2\% & 1.87& 1.04  & 70.2\% & 2.66 & 1.02  & 65.8\%  & 2.81 & 1.2 \\
Ours (V) & 72.5\%  & 2.37 & 0.98 & 72.5\%  & 2.37 & 0.98 & 72.5\%  & 2.37 & 0.98 \\
Ours (A+V) & \textbf{74.5\%} & 1.86 & 1.16  & \textbf{78.3\%} & 1.96 & 1.12 & \textbf{78.8\%}  & 1.81 & 1.18 \\
Ours (Unified) & 68.1\% & \textbf{1.68} & 1.92  & 75.6\%  & \textbf{1.49} & 1.76 & 75.9\% & \textbf{1.67} & 1.76  \\
\midrule
GT & - & 0  & 1.5 & - & 0 & 1.5 & - & 0 & 1.5 \\
\bottomrule
\end{tabular}
}
\label{tab:fto}
\vskip -0.05in
\end{table*}

\begin{table*}[!t]
\caption{Semantic evaluation of dialogue responses under different noise conditions.}
\vskip -0.1in
\centering
{\footnotesize
\begin{tabular}{lccccccc}
\toprule
Noise condition &  Clean&  & BG &  & Interf & \\
\midrule
Model & PPL($\downarrow$) & Pickup Ratio($\uparrow$) & PPL($\downarrow$) & Pickup Ratio($\uparrow$)  & PPL($\downarrow$) & Pickup Ratio($\uparrow$)  \\
\midrule
Moshi & 44.1 & 23.8\% & 52.4 & 19.4\% & 46.4 & 18.4\%\\
SE + Moshi & 50.8 & 26.4\% & 50.4 & 24.5\% & 56.1 & 19.1\%\\
Ours (ICL) & 25.8 & \textbf{66.6\%} & 24.6 & \textbf{68.1\%} & \textbf{23.1} & \textbf{67.8\%} \\
Ours (IT) & \textbf{23.2} & 32.5\% & \textbf{24.0} & 30.3\% & 23.8 & 36.7\%\\
Ours (Unified) & 31.0 & 29.6\% & 29.8 & 35.5\% & 32.5 & 31.3\% \\
\midrule
GT & 51.7 & - & 51.7 & - & 51.7 & - \\
\bottomrule
\end{tabular}
}
\label{tab:semantic}
\vskip -0.15in
\end{table*}

\subsection{Turn-Taking Prediction Evaluation}


We evaluate our model using turn-taking events from~\cite{nguyen2023generative} and measure floor-transfer offset (FTO), which is the duration between turn transitions, where negative FTO indicates overlap and positive FTO indicates a gap.


We extract the agent’s turn start timestamp using the SOT token and compute FTO as the gap between the user’s turn end (obtained from ground-truth turn  annotations) and the agent’s turn start. 

\noindent{\bf Metrics.}  We report three metrics:
\squishlist
\item  {\it Response Ratio:} percentage of FTOs within –2s to 3s, the typical range in InterAct's human conversations (around 90\% of FTOs in the InterAct test set are in this range).

\item {\it FTO Error:} mean absolute error (MAE) between generated and ground-truth FTOs.

\item {\it Median FTO:} the median value of FTOs.
\squishend


\noindent{\bf Baselines.} We compare with Moshi~\cite{moshi}, a state-of-art spoken dialogue model. 
\squishlist
\item \textbf{Moshi:} We deploy the Moshi checkpoints~\cite{moshi} to generate responses.

\item \textbf{SE+Moshi:} We first apply the speech enhancement (SE) model Demucs~\cite{defossez2020realtimespeechenhancement}, then feed the denoised audio to Moshi~\cite{moshi} to generate responses.

\squishend


\noindent{\bf Results.} As shown in Table~\ref{tab:fto}, the audio-visual dialogue model achieves the highest response ratio across all noisy scenarios: 74.5\% (\textit{Clean}), 78.3\% (\textit{BG}), and 78.8\% (\textit{Interf}), substantially outperforming the baseline Moshi model, which reaches only 50\%. Adding visual input to the audio-only model improves turn-taking accuracy by 1.3\% (Clean), 8.1\% (BG), and 13\% (Interf). The unified audio-visual model shows a slight drop in response ratio but achieves the lowest FTO errors. Note that the GT median FTO is around 1.5s for the InterAct conversation dataset which consists of casual conversations between strangers. The detailed FTO distribution visualization can be found in \xref{sec:fto}.

To assess robustness under noise, in Fig.~\ref{fig:snr}B, we also evaluate turn-taking prediction across different SNR ranges for both \textit{BG} and \textit{Interf} scenarios. 




\subsection{Semantic Evaluation}
We evaluate the semantic quality of dialogue responses, comparing the Moshi baselines with three audio-visual dialogue variants:

\squishlist

\item \textbf{Ours (ICL):} Dual-model pipeline using in-context learning with example conversations from InterAct in the prompt of the \textsc{LLaMA3-8B} text backbone model (see~\xref{sec:texttrain}). 



\item \textbf{Ours (IT):} Dual-model pipeline fine-tuned via instruction tuning on Fisher, and InterAct datasets for the  text backbone model (see~\xref{sec:B2}).


\item \textbf{Ours (Unified):} Unified model trained to generate text responses from audio-visual input.


\squishend

We run our models end-to-end and compute the perplexity (PPL) of agent-generated turns. To further assess text quality, we use the Prometheus~\cite{kim2023prometheus} LLM as an evaluator framework, performing relative/pairwise comparisons rather than absolute scoring, which better aligns with human judgment~\cite{kiritchenko2017best,liusie2023llm}. For each evaluation, the LLM compares the ground-truth InterAct response with the model-generated text. We compute the {\it Pickup Ratio} as the fraction of responses in which the LLM prefers the model-generated text over the ground truth (see details in~\xref{sec:appC}). 


As shown in Table~\ref{tab:semantic}, the baseline Moshi and SE+Moshi models achieve the lowest Pickup Ratio according to the LLM evaluator. The audio-visual dialogue model using in-context learning (ICL) achieves the highest Pickup Ratio among all methods. In contrast, the same dual-model pipeline fine-tuned via instruction tuning (IT) and the unified audio-visual model show a reduced Pickup Ratio of 30–40\%. This drop is likely because InterAct dialogues often contain casual, unpredictable conversation; fine-tuning the generation task on such data can degrade response quality. For perplexity (PPL), both the ICL and IT audio-visual models achieve the lowest values. Finally, the response quality for the unified model is worse than the cascaded model settings for our AV-dialogue models. This is line with recent observations in the related domain of speech-to-speech dialog models~\cite{nvidia}, where cascade model responses outperform  unified models. 




\subsection{Human Evaluation}

We conducted a human evaluation with 18 participants to assess the end-to-end performance of our audio-visual dialogue model. Model text outputs were converted to speech using the Moshi streaming TTS~\cite{moshi}, ensuring a fair comparison since both systems used the same TTS. Participants were given dialogue transcripts and audio, including both user turns and model responses.


We randomly selected 15 samples from the InterAct test set across Clean, BG, and Interf conditions (details and SNRs in~\xref{sec:app:sampledistribution}). For each sample, participants evaluated four conditions: (1) SE+Moshi, (2) dual-model + ICL, (3) unified model, and (4) ground truth. Each participant rated 8 dialogue sets, with randomized method order to avoid bias. Ratings followed the Mean Opinion Score (MOS) protocol (ITU-T P.808~\cite{ITU-P808-2018}) on a 5-point Likert scale, evaluating Naturalness (N-MOS) and Helpfulness (H-MOS) (see~\xref{sec:app:scales}).

Table~\ref{tab:human} shows that both our dual and unified models outperform the SE+Moshi baseline in naturalness and helpfulness. The dual model with in-context learning achieves the best results. The unified model drops by 0.5 in naturalness and 1.02 in helpfulness compared to the dual model. This is likely due to (1) the limited size of real conversational data and (2)  the fact that the conversations in the dataset are mostly casual chit-chat, often containing low-quality responses, random topic shifts, and limited logical reasoning.

These results highlight that the dual model benefits from using real-world data primarily for turn-taking modeling while leveraging the pretrained text backbone for stronger generation quality. Notably, the MOS trends align with the LLM evaluator pick ratios in our semantic evaluation.


\begin{table}[t!]
\vskip -0.15in
\caption{Human evaluation of dialogue generation.  N-MOS: Mean Opinon Score on Naturalness of response. H-MOS: Mean Opinion Score on Helpfulness of response. }
\vskip -0.1in
\centering
{\footnotesize
\begin{tabular}{lccc}
\toprule
Model & N-MOS($\uparrow$) & H-MOS($\uparrow$) \\
\midrule
SE + Moshi & 2.39 & 2.10 \\
Ours (Dual+ICL) & \textbf{4.14} & \textbf{4.09} \\
Ours (Unified) & 3.54  & 3.02 \\
\midrule
GT & 3.92 & 3.62 \\
\bottomrule
\end{tabular}
}
\label{tab:human}
\vskip -0.15in
\end{table}

\begin{table}[b]
\caption{Acoustic \& semantic token comparison.}
\vskip -0.1in
\centering
{\footnotesize
\begin{tabular}{lcccc}
\toprule
AVSR WER(\%) $\downarrow$ & Clean & BG & Interf \\
\midrule
DinoSR (A) &  24.9 & 89.0 & 239.2  \\
DinoSR (A+V) & 26.9 & 83.0 & 67.0 \\
Acoustic (A) & 28.6 & 60.0 & 63.4 \\
Acoustic (A+V) & 16.3 & 37.4 & 30.8 \\
\midrule
Response Ratio(\%) $\uparrow$ & Clean & BG & Interf \\
\midrule
DinoSR (A) & 67.3 & 63.3 & 63.2 \\
DinoSR (A+V) & 69.5  & 49.1 & 47.8 \\
Acoustic (A) & 73.2 & 70.2 & 65.8 \\
Acoustic (A+V) & 74.5  & 76.9 & 78.8 \\
\bottomrule
\end{tabular}
}
\label{tab:dinosr}
\end{table}

\subsection{Ablation studies}

We first compare acoustic tokens with semantic tokens. Using the same training setup, we trained the AV model with DinoSR~\cite{liu2023dinosr} semantic tokens instead of DAC tokens, as DinoSR is a newer, improved semantic representation compared to HuBERT. Table~\ref{tab:dinosr} shows that acoustic tokens outperform semantic tokens in both streaming AVSR and turn-taking prediction tasks.



Next, we compare different training strategies (Table~\ref{tab:train}). {\it No Stage 1 } indicates skipping Stage 1 while training on the  LLaMA3-8B model, while {\it No Audio Dialogue} excludes the audio-only dialogue dataset during Stage 2 fine-tuning. Results show a significant drop without Stage 1. Including the audio-only dialogue dataset  improves  performance.


\begin{table}[t!]
\caption{Ablation Study on the training recipe.} 
\vskip -0.1in
\centering
{\footnotesize
\begin{tabular}{lcccc}
\toprule
AVSR WER(\%) $\downarrow$ & Clean & BG & Interf \\
\midrule
Ours(A+V) & 16.3 & 37.4 & 30.8  \\
Ours(No Stage 1) & 58.9 & 95.1 & 86.7 \\
Ours(No Audio Diag) & 22.6 & 37.8 & 31.8 \\
\bottomrule
\end{tabular}
}
\label{tab:train}
\vskip -0.15in
\end{table}

\begin{table}[t!]
\vskip 0.07in
\caption{Ablation study on  turn-taking supervision in our Unified Model.}
\vskip -0.1in
\centering
{\footnotesize
\begin{tabular}{lcccc}
\toprule
Response ratio(\%) $\uparrow$ & Clean & BG & Interf \\
\midrule
w explicit turn-taking  & 68.1 & 75.6 & 75.9  \\
w/o explicit turn-change  & 48.0 & 35.1 & 38.0 \\
\midrule
\multicolumn{2}{l}{LLM-evaluator pick-up ratio(\%) $\uparrow$ } & & \\
\midrule
w explicit turn-change  & 29.6 & 35.5 & 31.3  \\
w/o explicit turn-change  & 29.0 & 22.1 & 18.2 \\

\bottomrule
\end{tabular}
}
\label{tab:turntaking}
\vskip -0.15in
\end{table}

We also  evaluate the impact of explicit turn-taking supervision on the unified model. In Stage 2, we trained a version without the \verb|<SOT>| token, forcing it to generate response tokens directly. Table~\ref{tab:turntaking} shows that this supervision is crucial for both turn-taking prediction and response quality.


%% file: conclude-1.tex
\section{Conclusion}

We introduced AV-Dialog, the first  streaming audio-visual dialogue system that integrates audio, vision, turn-taking, and response generation. Using acoustic tokenization, multi-stage training, and explicit turn-event supervision, it achieves robust performance in noisy, multi-speaker environments, outperforming audio-only baselines in transcription, turn-taking, and response quality. Human evaluations confirm that AV-Dialog enables more natural, helpful, and speaker-aware conversations, underscoring the value of combining listening and looking for real-world multimodal dialogue.

%% file: limits-1.tex
\section{Limitations and Risks}

\noindent\textbf{Limitations.} While AV-Dialog advances full-duplex dialogue in noisy environments, its performance can be further improved. It currently does not explicitly model non-verbal auditory cues (e.g., laughter, sighs) or visual cues (e.g., facial expressions, gestures) beyond lip movements. Enhancing the understanding and generation of these multimodal signals could make interactions more human-like. Finally, factors like poor lighting, occlusions (e.g., hands covering the mouth) or extreme head poses,  can impair lip movement extraction~\cite{avhubert}, affecting speaker tracking and speech understanding. Developing lip encoders that are robust to such conditions is a promising and complementary direction for future work.


\noindent\textbf{Ethical considerations.} Like any advanced AI enabling human-like interaction, AV-Dialog presents key ethical challenges. It may produce  misleading dialogue, particularly under noisy or ambiguous conditions, requiring rigorous evaluation and ongoing monitoring. While audio-visual  capture (e.g., lip movements, voices) is common in voice conferencing platforms like Zoom, it still demands strict attention to privacy. To prevent misuse such as exploitation in online scams, methods like speech watermarking could help safeguard against abuse.